\documentclass{article}
\usepackage{adjustbox}
\usepackage{algorithm}
\usepackage{algorithmic}
\usepackage{amsfonts}
\usepackage{amsmath}
\usepackage{array}
\usepackage{babel}
\usepackage{bbold}
\usepackage{bm}
\usepackage{booktabs}
\usepackage{diagbox}
\usepackage{graphicx}
\usepackage{inputenc}
\usepackage{makecell}
\usepackage{microtype}
\usepackage{multirow}
\usepackage{nicefrac}
\usepackage{pifont}
\usepackage{spconf}
\usepackage{stfloats}
\usepackage{textcomp}
\usepackage{ulem}
\usepackage{url}
\usepackage{verbatim}
\usepackage{wrapfig}
\usepackage{xcolor}
\usepackage{subfigure}

\usepackage[
bookmarks=true,
colorlinks=true,
linkcolor=blue,
urlcolor=black,
citecolor=blue,
bookmarks=true,
linktocpage=true,   
hyperindex=true
]{hyperref}

\title{
Time-independent Spiking Neuron via Membrane Potential Estimation for Efficient Spiking Neural Networks}

\name{Hanqi Chen$^{1}$, Lixing Yu$^{1*}$, Shaojie Zhan$^{1}$, Penghui Yao$^{1}$, Jiankun Shao$^{2}$\thanks{* Corresponding author. The work was supported in part by The National Natural Science Foundation of China (No.62106217), Yunnan Provincial Department of Science and Technology Basic Research Program (No.202201AU070112) and the Open Project Program of Yunnan Key Laboratory of Intelligent Systems and Computing (No.202405AV340009).}}
\address{
  $^{1}$School of Information Science and Engineering, Yunnan University, \\ Yunnan Key Laboratory of Intelligent Systems and Computing, Yunnan, China \\
  $^{2}$State Key Laboratory of Explosion Science and Technology, Beijing Institute of Technology, \\ Beijing, China
}

\begin{document}
%
\maketitle
\begin{abstract}
The computational inefficiency of spiking neural networks (SNNs) is primarily due to the sequential updates of membrane potential, which becomes more pronounced during extended encoding periods compared to artificial neural networks (ANNs). This highlights the need to parallelize SNN computations effectively to leverage available hardware parallelism. To address this, we propose Membrane Potential Estimation Parallel Spiking Neurons (MPE-PSN), a parallel computation method for spiking neurons that enhances computational efficiency by enabling parallel processing while preserving the intrinsic dynamic characteristics of SNNs. Our approach exhibits promise for enhancing computational efficiency, particularly under conditions of elevated neuron density. Empirical experiments demonstrate that our method achieves state-of-the-art (SOTA) accuracy and efficiency on neuromorphic datasets. Codes are available at~\url{https://github.com/chrazqee/MPE-PSN}.
\end{abstract}

\begin{keywords}
Parallel Spiking Neurons, Membrane Potential Estimation, Efficient Computing
\end{keywords}

\section{Introduction} \label{sec:intro}
At the heart of spiking neural networks (SNNs) lie dynamic spiking neurons, intricately handling inputs, revealing intricate neuronal dynamics, and triggering spikes upon membrane potential thresholds being surpassed.
While spiking neurons currently operate with sequential computational updates, their energy efficiency coexists with computational inefficiency, leading to latency issues in training and inference compared to  artificial neural networks (ANNs).

Prior advancements in spiking neurons, such as PLIF~\cite{plif} integrating a learnable time constant, IM-LIF~\cite{IM-LIF} incorporating attention mechanisms, GLIF~\cite{glif} featuring gating mechanisms, and KLIF~\cite{klif} optimizing activation slopes and surrogate gradient curve width, have predominantly operated within a sequential computational framework. This underscores the need for exploring parallelized computation approaches leveraging the potential of spiking neurons.

Research by Parallel Spiking Neurons (PSN)~\cite{psn} introduces a parallelized spiking neuron design by circumventing the reset process inherent in SNNs. However, this approach compromises the distinctive high dynamic properties of spiking neurons, causing the operational mechanism of SNNs to mirror that of ANNs instead. And Parallel Multi-compartment Spiking Neuron (PMSN)~\cite{pmsn} elevates neuronal dynamics and enables the parallelization of dynamic neuronal computations, albeit maintaining temporal dependence.

\begin{figure}[t]
  \centering
  \includegraphics[width=0.8\linewidth]{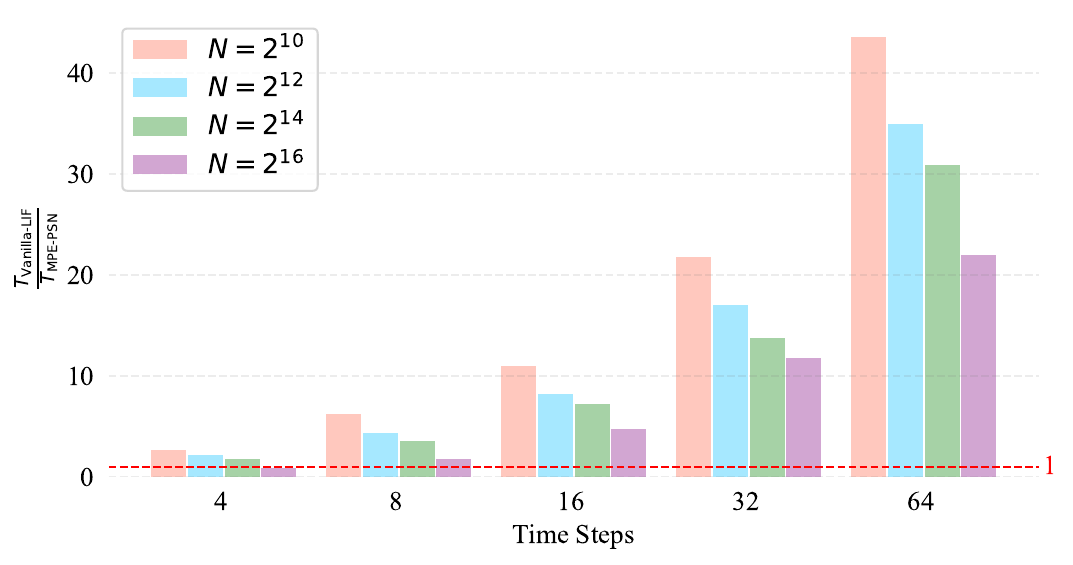}
  \vspace{-0.3cm}
  \caption{
  Variations in the execution time ratio between the vanilla-LIF neuron and the MPE-PSN neuron concerning time step increments and neuron quantity augmentation in both cases during the forward process, denoted as $\frac{T_{vanilla-LIF}}{T_{MPE-PSN}}$.}
  \label{fig: speed ratio}
  \vspace{-0.1 in}
\end{figure}
To address prevailing challenges, we introduce a Membrane Potential Estimation Parallel Spiking Neuron (MPE-PSN), a highly dynamic and fully parallelized computational neuron. 
MPE-PSN estimates the activation output through the spiking probability of the membrane potential
, thereby decoupling spiking neuron computations from temporal dependencies. This strategy promises a substantial boost in the computational efficiency of long-temporal dependent spiking neural networks, as exemplified by~\cite{spikingvit}. Throughout SNN training, the alignment between true and estimated membrane potentials is optimized using the minimum mean square error (MSE) loss function. The empirical findings validate the marked improvement in computational efficiency offered by the proposed methodology while preserving the intricate dynamic attributes of spiking neurons. Notable performance benchmarks are attained on neuromorphic datasets.


\section{Methods} \label{sec:methods}

\begin{figure*}[t]
  \centering
  \includegraphics[width=0.95\linewidth]{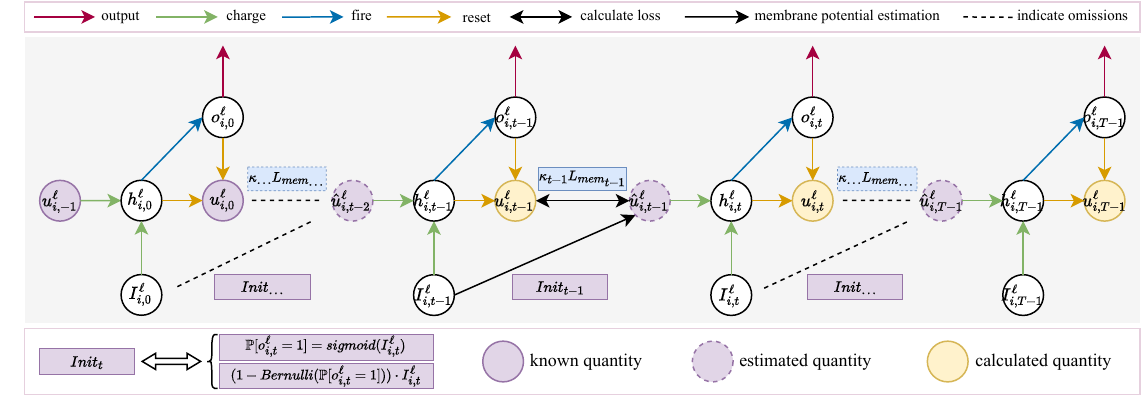}
  \vspace{-0.3cm}
  \caption{Schematic diagram of the proposed spiking neuron internal structure based on membrane potential estimation.} 
  \label{fig: main}
  \vspace{-0.05 in}
\end{figure*}

\subsection{Motivation and Derivation of Neuron Design} \label{motivation}
Spiking neurons perform efficiently in temporal information processing.
Their iterative computational process can be outlined as follows:
\begin{align}
    h_{i,t}^{\ell} &= g(u_{i,t-1}^{\ell}, I_{i,t}^{\ell}), \label{neuron_charge} \\
    o_{i, t}^{\ell} &= \Theta(h_{i,t}^{\ell} - v_{th}) = \left\{\begin{array}{l}
                                                        1, ~~ h_{i,t}^{\ell} \geq v_{th}, \\
                                                        0, ~~ otherwise,
                                                        \end{array}\right. \label{neuron_fire} \\
    u_{i,t}^{\ell} &= h_{i,t}^{\ell} \cdot (1 - o_{i,t}^{\ell}) + v_r \cdot o_{i,t}^{\ell}, \label{neuron_reset}
\end{align}

\noindent where $I_{i,t}^{\ell}$ represents the input current of the $i$-th neuron in the $l$-th layer at timestep $t$, $u_{i,t}^{\ell}$ represents the membrane potential post-spike firing and reset, and $h_{i,t}^{\ell}$ represents the integrated membrane potential prior to firing. The threshold for membrane potential is denoted as $v_{th}$, and $o_{i, t}^{\ell}$ represents the output following spike firing. The function $g(\cdot)$ in Equation~(\ref{neuron_charge}) corresponds to different types of spiking neurons, while $\Theta(\cdot)$ in Equation~(\ref{neuron_fire}) represents the Heaviside step function. In Equation~(\ref{neuron_reset}), $v_r$ stands for the reset potential. This equation elucidates a hard reset scenario in a spiking neuron, where the membrane potential is reset from $u_{i,t}^{\ell}$ to $v_r$ upon firing. In surrogate training, the hard reset method is favored over the soft reset approach for enhanced performance~\cite{Ledinauskas2020TrainingDS}.

By extending the Equations~(\ref{neuron_charge}) to~(\ref{neuron_reset}) to Leaky Integrate-and-Fire (LIF) neuron~\cite{Direct_Training}, i.e. $h_{i,t}^\ell = \tau_mu_{i,t-1}^\ell + I_t$, we obtain the Equation~(\ref{expand}) as:
\begin{align}
    u_{i,t}^{\ell} = (\tau_m u_{i,t-1}^{\ell} + I_{i,t}^{\ell})(1-\Theta(\tau_m u_{i,t-1}^{\ell} + I_{i,t}^{\ell} - v_{th})), \label{expand}
\end{align}
where $\tau_m$ represents the membrane potential constant, input decay is disregarded, and $v_r$ is set to 0 in this work, the iterative update of a spiking neuron adheres to a sequential, incremental process. 
At time step $t$, the output is contingent on the preceding time step $t-1$. The computational time complexity of the neuron at this stage is $O(T)$, with $T$ denoting the total number of time steps.
Given the assumption of independence of membrane potential updates from the previous time point's output, spiking neuron updates can be architected for parallel processing.


Apparently, when $t = 0$, the previous time point's membrane potential of the neuron is noted as 0, i.e. $u_{i,-1}^{\ell} = 0$. Consequently, the computation for $u_{i,0}^{\ell}$ is exact. Referring to Equation~(\ref{expand}), we obtain:
\begin{align}
    u_{i,0}^{\ell} = I_{i,0}^{\ell}(1-\Theta(I_{i,0}^{\ell} - v_{th})),
\end{align}

\noindent Subsequently, we initialize the estimated values of $u_{i,1}^{\ell}$, referred to as $\hat{u}_{i,1}^{\ell}$, to the
estimated values of $u_{i,T-1}^{\ell}$, denoted as $\hat{u}_{i,T-1}^{\ell}$, which is determined by $I_{i,t}^{\ell}$, calculated as:
\begin{align}
    I_{i,t}^{\ell} & = \sum_{j=1}^{N_{\ell-1}}w_{i,j}^{\ell-1}o_{j,t-1}^{\ell-1},
\end{align}
where $w_{i,j}^{\ell-1}$ represents the synaptic weight between the $j$-th neuron in the preceding layer and the $i$-th neuron in the subsequent layer. $N_{\ell-1}$ denotes the number of neurons in the $(\ell-1)$-th layer. 

The accurate $u_{i,0}^{\ell}$ is utilized to rectify the error, denoted as $\mathcal{E}$. 
The procedure is outlined as follows:
\begin{align}
    u_{i,t}^{\ell} &= (\tau_m \hat{u}_{i,t-1}^{\ell} + I_{i,t}^{\ell})(1-\Theta(\tau_m \hat{u}_{i,t-1}^{\ell}+I_{i,t}^{\ell} - v_{th})), \\
    u_{i,t+1}^{\ell} &= (\tau_m \hat{u}_{i,t}^{\ell} + I_{i,t+1}^{\ell})(1-\Theta(\tau_m \hat{u}_{i,t}^{\ell}+I_{i,t+1}^{\ell} - v_{th})), \\
    \mathcal{E} &= error(\hat{u}_{i,t}^{\ell}, u_{i,t}^{\ell}). \label{error}
\end{align}
The comprehensive illustration of the entire process is shown in Fig.~\ref{fig: main}. 

\subsection{Initialization of Estimated Membrane Potential}
The output of spiking neurons adheres to the $Bernoulli$ distribution.
According to the relevant definition in~\cite{MA2023100831}, this is described as:
\begin{align}
    o_{i,t}^{\ell} \sim Bernoulli(\mathbb{P}[o_{i,t}^{\ell}=1]),
\end{align}
\noindent where $\mathbb{P}[o_{i,t}^\ell = 1]$ represents the probability that $o_{i,t}^\ell = 1$. In our study, we employed the $sigmoid$ function and the input current $I_{i,t}^\ell$ to compute $\mathbb{P}[o_{i,t}^\ell = 1]$, thereby evaluating the firing probabilities of each neuron. Subsequently, the computed probability is utilized to estimate the membrane potential. The calculation procedures in matrix format are outlined as follows:
\begin{align}
    \mathbb{P}[\mathbf{o}=\mathbf{1}] &= sigmoid(\mathbf{I}), ~\mathbf{I}\in \mathbb{R}^{T\times C \times H \times W}\\
    \mathbf{\hat{u}} &= (1-Bernulli(\mathbb{P}[\mathbf{o}=\mathbf{1}])) \odot \mathbf{I}.
\end{align}

\begin{table}[!t]
\centering
\caption{
Training specifics and network architectures for diverse datasets. SGD: Stochastic Gradient Descent. C: Channel. MP: Max Pooling. AP: Average Pooling. FC: Fully Connected.}
\vspace{0.1cm}
\label{tab:detials}
\resizebox{0.485\textwidth}{!}{%
\begin{tabular}{cccccccc}
\hline\hline
Dataset       & \begin{tabular}[c]{@{}c@{}}Learning\\ rate\end{tabular} & \begin{tabular}[c]{@{}c@{}}Opti-\\ mizer\end{tabular} & \begin{tabular}[c]{@{}c@{}}Batch\\ size\end{tabular} & Epoch & Resolution     & Network structure                                                                                            \\ \hline
DVSGesture    & 0.05                                                   & SGD                                                   & 8                                                    & 230   & $64\times64$   & 5$\times$(256C3-MP2)-FC-FC-AP                                                                                \\ \cline{7-7} 
CIFAR10-DVS   & 0.05                                                   & SGD                                                   & 32                                                   & 230   & $48\times48$   & \begin{tabular}[c]{@{}c@{}}64C3-128C3-AP2-256C3-256C3-AP2-\\ 512C3-512C3-AP2-512C3-512C3-AP2-FC\end{tabular} \\ \cline{7-7} 
N-Caltech 101 & 0.01                                                   & SGD                                                   & 8                                                    & 230   & $128\times128$ & \begin{tabular}[c]{@{}c@{}}64C3-128C3-AP2-256C3-256C3-AP2-\\ 512C3-512C3-AP2-512C3-512C3-AP2-FC\end{tabular} \\ \cline{7-7} 
CIFAR10 / 100   & 0.1                                                    & Adam~\cite{adam}                                      & 64                                                   & 300   & $32\times32$   & ResNet19~\cite{7780459}                                                                            \\ \hline\hline

\end{tabular}%
}
\vspace{-0.15 in}
\end{table}

\noindent where bold font denotes variables that are matrices. $T$ represents the time steps of the simulation, while $C$, $H$, and $W$ denote the channel, height, and width of the accumulated frames, respectively. The symbol $\odot$ represents the Hadamard product. The time complexity for computing $\mathbb{P}[\mathbf{o}=\mathbf{1}]$ is theoretically $O(1)$, as is that of the $Bernoulli$ operator. Consequently, the additional time required for this operation is negligible, particularly on graphics processing unit (GPU) devices.


\subsection{Membrane Potential Approximation Loss}
We have addressed the discrepancy between the probabilistically initialized estimate, $\hat{u}_{i,t}^{\ell}$, and the computationally obtained value $u_{i,t}^{\ell}$ by applying a correction mechanism (i.e., in Section~\ref{motivation}). Such disparity is mitigated through the utilization of the minimum mean square error (MSE) loss function to approximate the true and probabilistically initialized values. Crucially, only $u_{i,0}^{\ell}$ signifies the precise calculated arrival. Efficient approximation of both $\hat{u}_{i,t}^{\ell}$ and $u_{i,t}^{\ell}$ necessitates weighting different moments with distinct learnable coefficients $\kappa_i$. It formalizes the expression of $\mathcal{L}_{mem}$ as outlined:

\begin{align}
    \mathcal{L}_{mem} = \sum_{t=0}^{T-1}\sum_{i=0}^{N_\ell-1}\kappa_i MSE(\hat{u}_{i,t}^{\ell}, {u}_{i,t}^{\ell}),
\end{align}
The overall loss is calculated as:
\begin{align}
    \mathcal{L}_{total} = (1-\lambda)\mathcal{L}_{cls} + \lambda\mathcal{L}_{mem},
\end{align}
where $\lambda$ represents a weighting coefficient, set to 0.01. $\mathcal{L}_{\text{cls}}$ denotes the classification loss, which is computed employing the methodology outlined in the~\cite{deng2022temporal}. The loss function of $\mathcal{L}_{cls}$ is 
$
    \frac{1}{T}\sum_{t=0}^{T-1}\mathcal{L}_{ce}[\mathbf{o}_t, \mathbf{y}], 
$
where  $\mathbf{o}_t$, $\mathbf{y}$, and $\mathcal{L}_{ce}$ represent outputs, labels, and cross entropy loss respectively.

\section{Expriments} \label{sec:expriment}
\subsection{Experimental Details}
In direct training scenarios, the gradient is computed using the spatio-temporal backpropagation (STBP)~\cite{stbp} algorithm, as articulated:
\begin{align}
   \frac{\partial L}{\partial w_{i,j}^{\ell}} = \sum_{t=1}^{T} \frac{\partial L}{\partial o_{i,t}^{\ell}}\frac{\partial o_{i,t}^{\ell}}{\partial {u}_{i,t}^{\ell}}\frac{\partial {u}_{i,t}^{\ell}}{\partial I_{i,t}^{\ell}}\frac{\partial I_{i,t}^{\ell}}{\partial w_{i,j}^{\ell}},
\end{align}
where $\frac{\partial o_{i,t}^{\ell}}{\partial {u}_{i,t}^{l}}$ are non-differentiable, and gradients are typically approximated using integrable functions that share similar shapes. In our study, a triangular function~\cite{deng2022temporal} is utilized for gradient approximation, represented mathematically as:
\begin{align}
    \frac{\partial o_{i,t}^{\ell}}{\partial {u}_{i,t}^{\ell}} = \frac{1}{\alpha^2}max(0, \alpha - |{u}_{i,t}^{\ell}-v_{th}|).
\end{align}
here, $\alpha$ is a constant that restricts the range of activated gradients, fixed at 1.0. The hyper-parameter configurations and network architecture details are outlined in Table~\ref{tab:detials}. Additionally, $\tau_m$ is set to 0.25, with the threshold represented as a parameter subject to learning.

\subsection{Comparisons With Existing Methods}
We evaluated the MPE-PSN neuron in SNNs across both neuromorphic and static datasets, with the experimental results presented in Table~\ref{tab:main}.

\textbf{DVSGesture}~\cite{8100264}. 
We performed extensive classification experiments on the DVSGesture dataset using the 7B-Net architecture. Our method achieved a top-1 accuracy of 97.92\% at 16 time steps.
This outcome surpassed the SOTA result with 4 fewer time steps.

\begin{table}[!t]
    \caption{
    Compare with previous studies on neuromorphic and static datasets. T: time steps. B: learnable blocks. Prefix M signifies the term modified. Results marked with $\ast$ denote self-implemented findings. The threshold parameter for Vanilla LIF and PLIF is established at 1.0 for the N-Caltech 101 dataset.
    }
    \vspace{0.1cm}
    \label{tab:main}
    \resizebox{\linewidth}{!}{
    \begin{tabular}{ccccc}
    \hline\hline
    Dataset                                                                   & Approach                     & Network            & T                  & Accuracy(\%)                   \\ \hline\hline
    \multicolumn{1}{c|}{\multirow{5}{*}{\begin{tabular}[c]{@{}c@{}}DVS\\Gesture\end{tabular}}}     & KLIF ~\cite{klif}            & M-PLIF-Net         & 12         & 94.1              \\
    \multicolumn{1}{c|}{}                                                                          & MLF~\cite{ijcai2022p343}     & ResNet-20          & 40         & 97.29             \\
    \multicolumn{1}{c|}{}                                                                          & IM-LIF~\cite{IM-LIF}         & ResNet-19          & 40         & 97.33             \\
    \multicolumn{1}{c|}{}                                                                          & PLIF~\cite{plif}             & 7B-Net             & 20         & 97.57             \\ \cline{2-5} 
    \multicolumn{1}{c|}{}                                                                          & MPE-PSN                      & 7B-Net             & 16,20      & 97.92,97.22       \\ \hline\hline
    \multicolumn{1}{c|}{\multirow{5}{*}{\begin{tabular}[c]{@{}c@{}}CIFAR10\\ DVS\end{tabular}}}    & GLIF~\cite{glif}             & 7B-wideNet         & 16         & 78.1              \\
    \multicolumn{1}{c|}{}                                                                          & IM-LIF~\cite{IM-LIF}         & VGGSNN             & 10         & 80.5              \\
    \multicolumn{1}{c|}{}                                                                          & TET~\cite{deng2022temporal}  & VGGSNN             & 10         & 83.17             \\
    \multicolumn{1}{c|}{}                                                                          & SPSN~\cite{psn}              & VGGSNN             & 4,8,10     & 82.3,85.3,85.9    \\ \cline{2-5} 
    \multicolumn{1}{c|}{}                                                                          & MPE-PSN                      & VGGSNN             & 4,8,10     & 84.48, 86.09,86.69\\ \hline\hline
    \multicolumn{1}{c|}{\multirow{4}{*}{\begin{tabular}[c]{@{}c@{}}N-Caltech \\ 101\end{tabular}}} & Vanilla LIF                  & VGGSNN             & 10         & $87.13^\ast$      \\
    \multicolumn{1}{c|}{}                                                                          & PLIF~\cite{plif}             & VGGSNN             & 10         & $74.5^\ast$       \\
    \multicolumn{1}{c|}{}                                                                          & Mask-PSN~\cite{psn}          & VGGSNN             & 10         & $85.78^\ast$      \\ \cline{2-5} 
    \multicolumn{1}{c|}{}                                                                          & MPE-PSN                      & VGGSNN             & 10         & 88.24             \\ \hline\hline
    \multicolumn{1}{c|}{\multirow{5}{*}{\begin{tabular}[c]{@{}c@{}}CIFAR\\ 10\end{tabular}}}       & KLIF~\cite{klif}             & M-PLIF-Net         & 10         & 92.52             \\
    \multicolumn{1}{c|}{}                                                                          & PLIF~\cite{plif}             & PLIF Net           & 8          & 93.50             \\
    \multicolumn{1}{c|}{}                                                                          & IM-LIF~\cite{IM-LIF}         & ResNet-19          & 6          & 95.66             \\
    \multicolumn{1}{c|}{}                                                                          & PSN~\cite{psn}               & M-PLIF-Net         & 4          & 95.32             \\ \cline{2-5} 
    \multicolumn{1}{c|}{}                                                                          & MPE-PSN                      & ResNet-19          & 4          & 94.36             \\ \hline\hline
    \multicolumn{1}{c|}{\multirow{4}{*}{\begin{tabular}[c]{@{}c@{}}CIFAR\\ 100\end{tabular}}}      & TET~\cite{deng2022temporal}  & ResNet-19          & 6          & 74.72             \\
    \multicolumn{1}{c|}{}                                                                          & TEBN~\cite{duan2022temporal} & ResNet-19          & 6          & 76.41             \\
    \multicolumn{1}{c|}{}                                                                          & IM-LIF~\cite{IM-LIF}         & ResNet-19          & 6          & 77.42             \\ \cline{2-5} 
    \multicolumn{1}{c|}{}                                                                          & MPE-PSN                      & ResNet-19          & 4,6        & 75.9,77.02        \\ \hline\hline
    \end{tabular}
    }
    \vspace{-0.1in}
\end{table}

\textbf{CIFAR10DVS}~\cite{10.3389/fnins.2017.00309}. 
We assessed the effectiveness of our method on the CIFAR10DVS dataset across time steps of 8, 10, and 16. 
Our results consistently outperform those of the current SOTA methods.

\textbf{N-Caltech 101}~\cite{10.3389/fnins.2015.00437}. 
The efficacy of our method was further confirmed on the N-Caltech 101 dataset, with the performance of other methods reproduced for comparison. The outcomes revealed that our approach achieved the highest accuracy presently attainable on the N-Caltech 101 dataset. 

\textbf{CIFAR10 \& CIFAR100}~\cite{2009Learning}. Our method achieved top-1 accuracy rates of 94.36\% and 75.9\% on the CIFAR10 and CIFAR100 static datasets, respectively. It is important to note that the performance of our method on static datasets is not exceptional due to the high homogeneity of the data caused by directly stacking images to simulate time steps, which leads to inaccurate probabilistic estimates and, consequently, affects performance.

\subsection{Membrane Potential Approximation Analysis}
\begin{figure}[t]
\begin{center}
\subfigure[$L_2~norm(\hat{\mathbf{u}},\mathbf{u})$ and Acc.]{\includegraphics[width=0.45\linewidth,trim=0 0 0 0,clip]{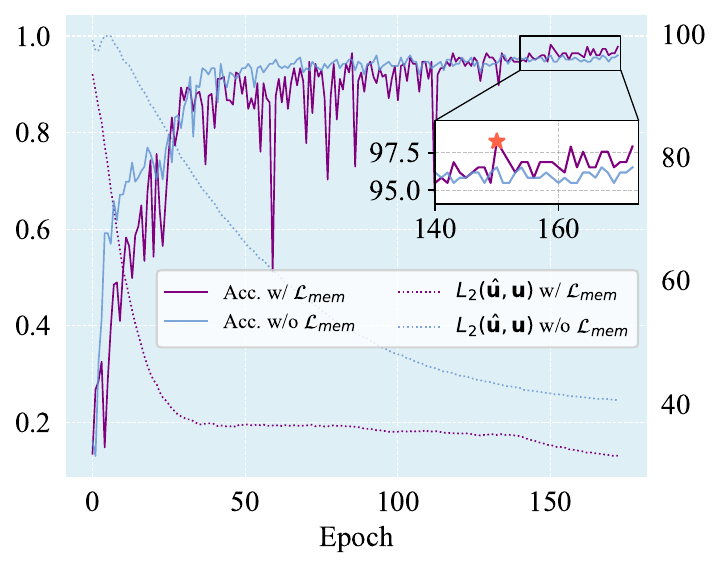}}
\subfigure[Spike rate (\%)]{\includegraphics[width=0.45\linewidth,trim=0 0 0 0,clip]{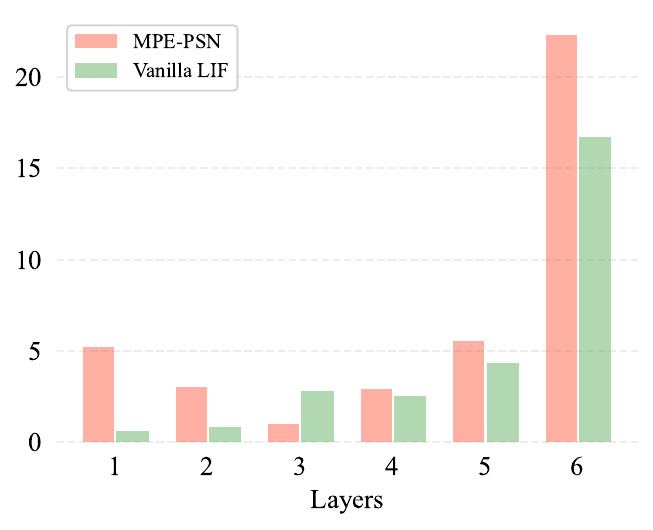}}
\vspace{-0.3cm}
\caption{(a): The curves of $L_2~norm$ between $\hat{\mathbf{u}}$ and $\mathbf{u}$, as well as the accuracy curves on the DVSGesture dataset. The values are normalised to a range of 0 to 1. (b): Spike rate of MPE-PSN neurons and vanilla LIF neurons on the DVSGesture dataset.} 
\label{fig: norm spike rate}
\end{center}
\vspace{-0.3in}
\end{figure}

We also make evaluation on the accuracy of the network model with and without $\mathcal{L}_{mem}$ and the impact of this loss on the $L_2$ norm of membrane potential distances. In Fig.~\ref{fig: norm spike rate} (a), for the network with $\mathcal{L}_{mem}$, initial epochs exhibit lower classification accuracy compared to the model without $\mathcal{L}_{mem}$ as the network emphasizes approximating membrane potential distances. Over epochs, the $\mathcal{L}_{mem}$-included model achieves superior accuracy, consistently maintaining a smaller $L_2$ norm of membrane potential distances than the model without $\mathcal{L}_{mem}$.

\begin{figure}[t]
\begin{center}
    \subfigure[Vanilla-LIF neuron]{\includegraphics[width=0.32\linewidth,trim=0 0 0 0,clip]{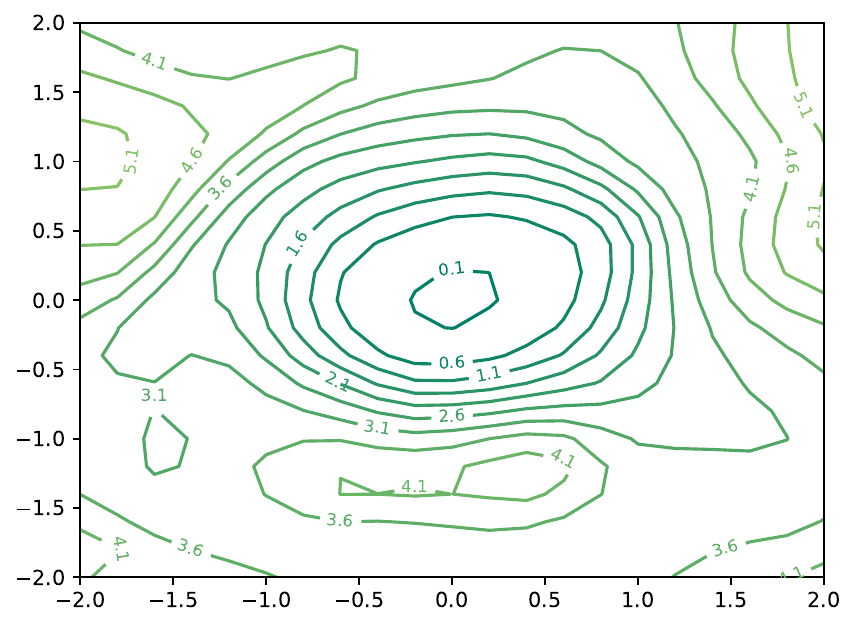}}
    \subfigure[MPE-PSN neuron]{\includegraphics[width=0.32\linewidth,trim=0 0 0 0,clip]{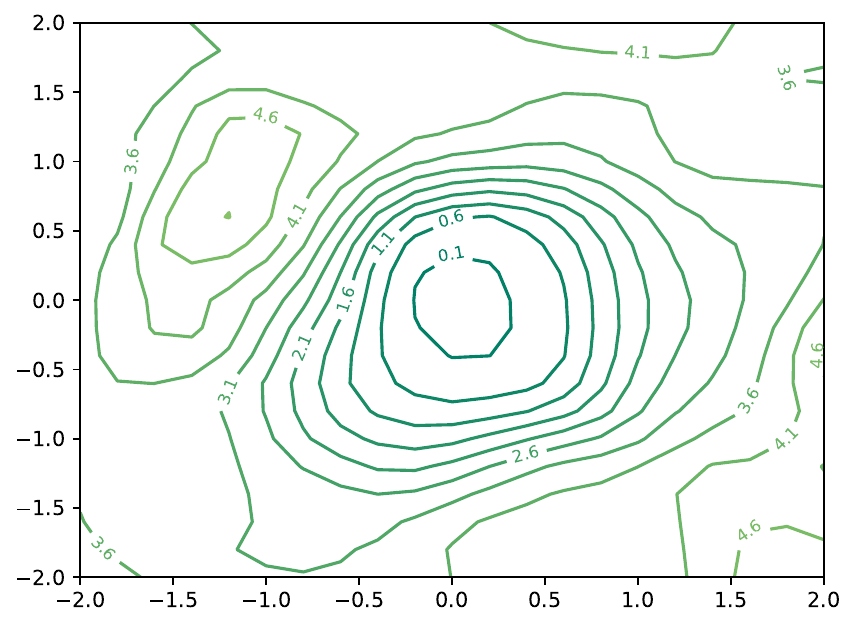}}
    \subfigure[Total Loss and Acc.]{\includegraphics[width=0.331\linewidth,trim=0 0 0 0,clip]{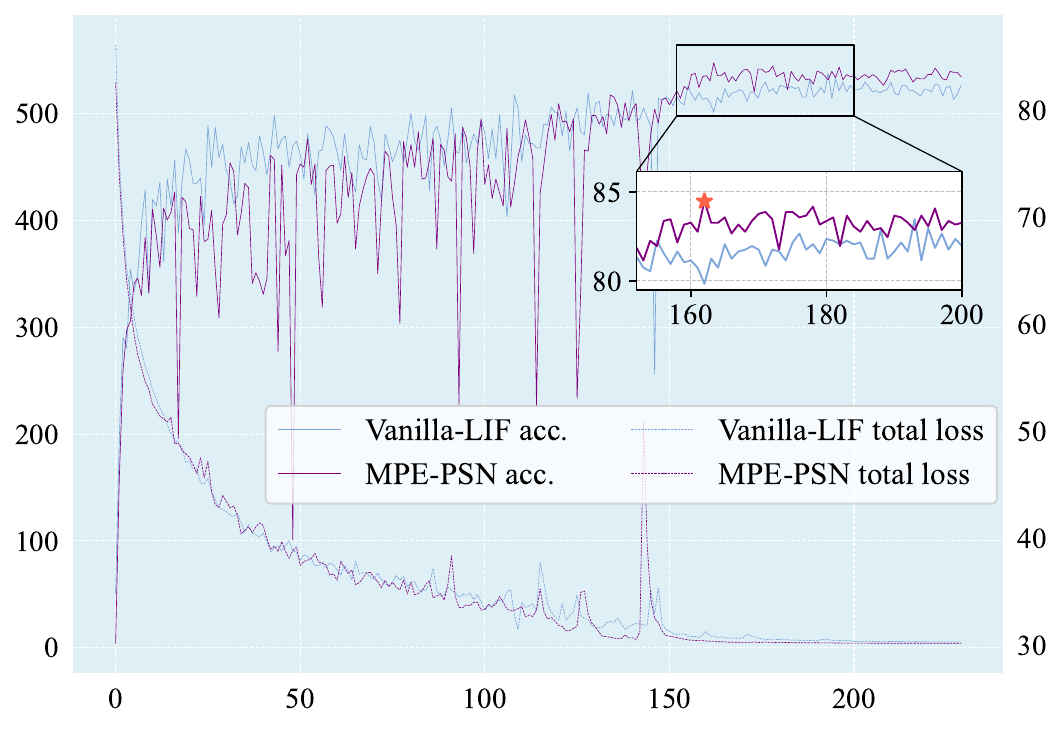}}
    \vspace{-0.3cm}
    \caption{
    2D loss contours, as introduced in \cite{visualloss}, along with accuracy and loss curves, are presented for the CIFAR10DVS dataset with a time step of 4.
    } 
    \label{fig: contour and acc loss curve}
\end{center}
\vspace{-0.3in}
\end{figure}

\subsection{Computational Efficiency Analysis}
A comparative analysis is conducted to assess the computational speed difference between our proposed neuron and the traditional LIF neuron across various time steps and neuron quantities. As depicted in Fig.~\ref{fig: speed ratio}, our neuron generally outpace vanilla LIF neuron in simulation speed. Furthermore, as shown in Fig.~\ref{fig: norm spike rate} (b), our proposed neuron exhibits a notably higher spike rate across most layers when utilizing probabilistic estimation, contrasting with the dynamic complexity observed in the vanilla LIF neuron.
Fig.~\ref{fig: contour and acc loss curve} (a) and \ref{fig: contour and acc loss curve} (b) illustrate that the loss contours of our proposed neuron are comparable to those of vanilla LIF neurons, with both displaying relatively flat profiles. Fig.~\ref{fig: norm spike rate} (a) and \ref{fig: contour and acc loss curve} (c) demonstrate that initial uncertainties in membrane potential estimation did impact the network’s fitting performance. However, as the estimation of membrane potential improved, this effect became less pronounced. Consequently, our proposed neuron maintain both computational efficiency and performance integrity.

\section{Conclusion} \label{sec:conclusion}
In this study, 
we introduce a parallel spiking neuron model centered on membrane potential estimation. This model enables computational parallelization while preserving neuronal dynamics, thereby notably improving the computational efficiency of spiking neural networks. Our neuron model has exhibited exceptional performance on neuromorphic datasets, pointing towards its capacity to drive the evolution of parallelized neurons in the future.





{\fontsize{9.5}{12}\selectfont
\bibliographystyle{IEEEbib}
\bibliography{refs}
}

\end{document}